\documentclass[11pt]{article}

\usepackage[final]{acl}

\usepackage{times}
\usepackage{latexsym}
\usepackage{amsmath}
\usepackage{amssymb}
\usepackage[table]{xcolor}
\usepackage{multirow}
\usepackage{booktabs}
\usepackage{array}
\usepackage{float}
\usepackage{adjustbox}
\usepackage[T1]{fontenc}
\usepackage[utf8]{inputenc}
\usepackage{textgreek}
\usepackage{xspace}
\usepackage{microtype}
\usepackage{inconsolata}
\usepackage{graphicx}
\usepackage{url}

\usepackage{framed}
\usepackage{enumitem}
\usepackage{listings}

\newcommand{\framework}{Structured On-policy Pruning}
\newcommand{\frameworkabbr}{STOP\xspace}
\newcommand{\ECN}{ECN\xspace}

\title{STOP: Structured On-Policy Pruning of Long-Form Reasoning in Low-Data Regimes}

\author{Chenjun Xu \\
  University of Washington\\
  \texttt{Chenjux@uw.edu} \\\And
  Zhennan Zhou \\
  University of Washington \\
  \texttt{zhouz46@cs.washington.edu} \\\And
  Zhan Su \\
  Halmstad University \\
  \texttt{zhan.su@hh.se} \\\AND
  Bill Howe \\
  University of Washington \\
  \texttt{billhowe@uw.edu} \\\And
  Lucy Lu Wang \\
  University of Washington \\
  \texttt{lucylw@uw.edu} \\\And
  Bingbing Wen \\
  University of Washington \\
  \texttt{bingbw@uw.edu} }

\begin{document}
\maketitle
\begin{abstract}
Long chain-of-thought (Long CoT) reasoning improves performance on multi-step problems, but it also induces \emph{overthinking}. This inefficiency is especially problematic in \emph{low-data} fine-tuning regimes, where real applications adapt reasoning models with limited supervision and cannot rely on large-scale teacher distillation or heavy test-time control. To address this, we propose \frameworkabbr\ (\framework), an on-policy algorithm for analyzing and pruning long-form reasoning traces. \frameworkabbr\ constructs self-distilled traces from the model. Then it maps each trace into a structured reasoning interface through node segmentation, taxonomy annotation, and reasoning-tree construction. On top of this interface, we introduce \ECN\ (Earliest Correct Node), which retains the shortest prefix ending at the earliest node. Experiments on DeepSeek-R1-Distill-Qwen-7B and DeepSeek-R1-Distill-LLaMA-3-8B across GSM8K, Math~500, and AIME~2024 show that \frameworkabbr\ reduces generated tokens by 19.4--42.4\% while largely preserving accuracy in low-data fine-tuning. Beyond efficiency, our analyses show that STOP induces much smaller distributional shift than teacher-guided pruning, improves the structural efficiency of generated reasoning, and reallocates reasoning effort away from redundant verification and backtracking toward more productive exploration.\footnote{Our code is available at:
\url{https://github.com/chenjux/ECN-STOP}.} 

\end{abstract}

\section{Introduction}
\begin{figure}[!t]
    \centering
    \includegraphics[width=\columnwidth]{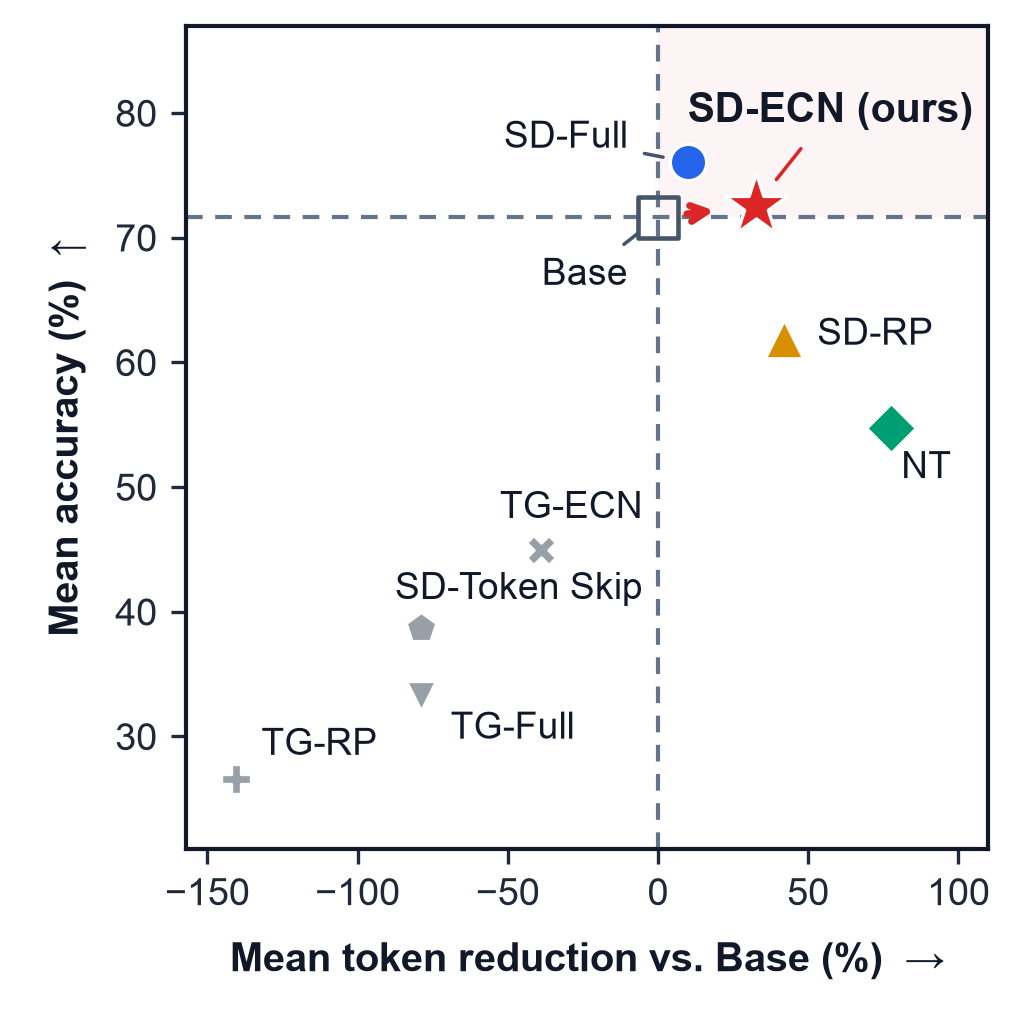}
    \caption{
    \textbf{Aggregate accuracy--token trade-off.}
    Using the Gemini-era results, SD-ECN reaches 72.6\%
    mean accuracy, 1.0 percentage point above Base (71.6\%),
    together with a 32.5\% mean token reduction relative to Base.
    SD-Full has the highest mean accuracy (76.1\%) but a
    smaller mean token reduction (9.9\%).
    Results are equally averaged over two backbones and three
    benchmarks; token reductions are computed relative to the
    corresponding Base before averaging.
    Dashed lines indicate Base; upper right is better.
    }
    \label{fig:overview}
\end{figure}
Reasoning language models (RLMs) have recently demonstrated strong
performance on complex multi-step tasks by generating long
chain-of-thought (Long CoT) traces~\citep{wei2022chain, yao2024tree,
jaech2024openai, guo2025deepseek}. However, longer reasoning does not
simply add useful intermediate steps---models often keep spending
computation after the answer is already available, a phenomenon known
as \emph{overthinking}~\citep{chen2024not, xu-etal-2025-language}. Such post-solution
reasoning inflates token usage, increases latency, and raises
deployment cost. The issue is especially acute for distilled reasoning
models, where efficiency is often a primary motivation for deployment,
and in low-data fine-tuning settings, where real applications have
limited supervision and cannot afford large-scale distillation or
expensive test-time control.

A growing body of work seeks to make long-form reasoning more efficient \citep{yang2025think,yi2025shorterbetter}, but existing approaches involve important trade-offs. Compression-based methods~\citep{ma2025cotvalve} rewrite or compress CoT into shorter representations, which can reduce length but may blur or remove critical intermediate logic. Pruning-based methods~\citep{zhao2025can} remove selected reasoning segments, but aggressive deletion can break logical continuity or suppress beneficial self-correction. Token-budget methods~\citep{han2025token} explicitly constrain reasoning length, but require manually chosen budgets and may under-allocate computation on genuinely difficult problems. Moreover, many efficient-reasoning methods~\citep{lippmann2025style} rely on externally rewritten or teacher-generated traces. In low-data regimes, such supervision can be poorly matched to the student’s native reasoning style, leading to substantial distributional shift during fine-tuning~\citep{chen2025retaining}.

In this work, we propose \frameworkabbr\ (\framework), an on-policy framework for structured analysis and pruning of long reasoning traces. \frameworkabbr\ has three tightly connected layers. First, it constructs supervision from \emph{self-distilled} successful traces generated by the same student model that will later be fine-tuned. This keeps training trajectories closer to the student’s native reasoning distribution. Second, it maps each trace into a \emph{structured reasoning interface} through node segmentation, taxonomy annotation, and reasoning-tree construction. The resulting representation decomposes a trace into coherent reasoning units, labels their functional roles (e.g., clarification, exploration, verification, backtracking, conclusion), and provides a structural view of how the reasoning unfolds. Third, built on top of this interface, we introduce \ECN\ (Earliest Correct Conclusion Node), a node-level pruning strategy that keeps only the minimal prefix up to the earliest node that already yields the correct answering conclusion.

Empirically, \frameworkabbr\ yields significant efficiency gains across GSM8K, Math~500, and AIME~2024. Across DeepSeek-R1-Distill-Qwen-7B and DeepSeek-R1-Distill-LLaMA-3-8B, STOP reduces generated tokens by 19.4--42.4\% while maintaining competitive accuracy. The results also reveal three broader patterns. First, self-distilled supervision is significantly more stable than teacher-guided supervision in the low-data regime, suggesting that on-policy trace construction is critical for preserving reasoning behavior. Second, node-level analysis shows that ECN does not merely truncate outputs: it systematically reduces redundant verification and backtracking while reallocating more of the reasoning budget toward exploration. Third, structural analysis of induced reasoning trees shows that STOP produces more focused and efficient reasoning morphologies, reducing redundant branching while largely preserving the core solution path. These findings suggest that effective pruning should be understood not only as length reduction, but as reorganization of the structure of generated reasoning.

\begin{figure*}[t]
    \centering
    \includegraphics[width=\linewidth]{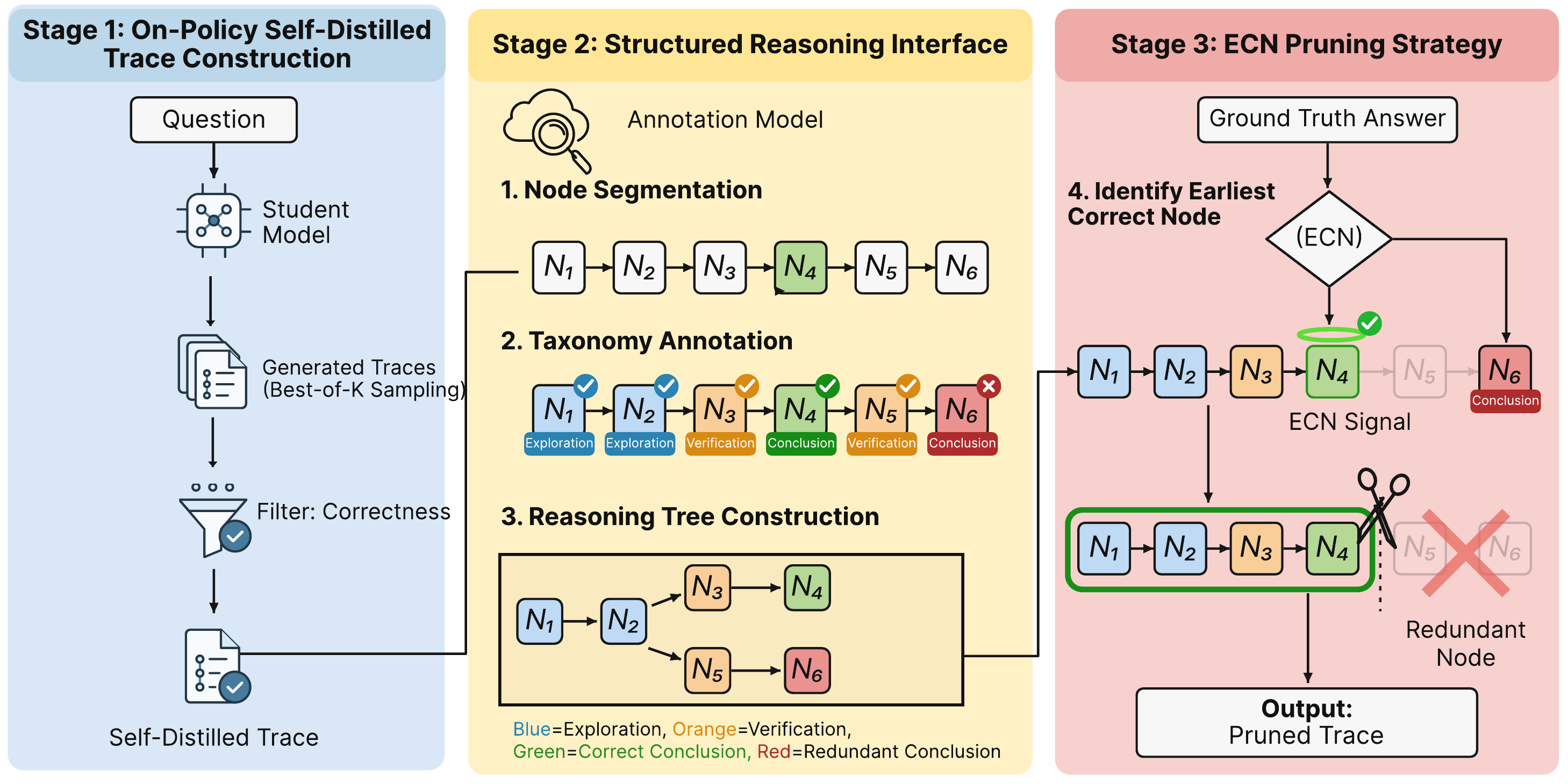} 
    \caption{Overview of \frameworkabbr. The framework first constructs successful
self-distilled traces, then maps each trace into a structured
reasoning interface through node segmentation, taxonomy annotation,
and reasoning-tree construction. On top of this interface, it applies
ECN pruning and supports analysis of reasoning behavior changes. }
    \label{fig:pipeline}
\end{figure*}

% \vspace{0.5em}
In summary, our contributions are:
\begin{itemize}[noitemsep, topsep=0pt, leftmargin=10pt]
  \item We introduce \frameworkabbr, a unified on-policy framework for long-form reasoning that combines self-distilled trace construction, a structured reasoning interface based on node segmentation / taxonomy annotation / tree construction, and node-level pruning.
    \item We propose \ECN\ (Earliest Correct Node), a pruning strategy defined on top of the structured node interface, which retains the minimal prefix up to the earliest correct answering conclusion while preserving semantic continuity. ECN consistently outperforms existing baselines. 
    \item We show that the same structured interface can be used to analyze reasoning behavior before and after pruning, revealing that our method does not merely shorten traces but shifts reasoning away from redundant verification and backtracking toward more productive exploration and more structurally efficient reasoning.
    \item We demonstrate that, in a low-data regime, \frameworkabbr\ achieves large token reductions with competitive accuracy, and that self-distilled supervision is substantially more stable than teacher-guided supervision for fine-tuning distilled reasoning models.
\end{itemize}

% \begin{figure}[t]
%     \centering
%     \includegraphics[width=0.5\linewidth]{fig/fig1-1.pdf} 
%   \caption{STOP with ECN reduces overthinking: compared to the Base model’s repetitive verification cycles, the fine-tuned model reaches the same correct answer with a shorter, more direct trace. \todoit{This figure needs to be improved， maybe integrate figure 1 into figure 2}}

%     \vspace{-5mm}
%     \label{fig:fig1}
% \end{figure}

% \vspace{-2mm}

\section{Method}
\label{sec:method}

We introduce the details of \frameworkabbr\ (\framework).  As shown in Figure~\ref{fig:pipeline}. \frameworkabbr\ has three tightly connected components: (i) self-distilled trace construction; (ii) Structured reasoning interface; and (iii) \ECN\ (Earliest Correct Node) pruning.

\subsection{On-Policy Self-Distilled Trace Construction}

The first component of \frameworkabbr\ is \textbf{self-distilled trace construction}. Here, \emph{self-distilled} means that the supervision traces are generated by the same student model that is later fine-tuned. This keeps the supervision distribution close to the student's native reasoning behavior and reduces the mismatch introduced by teacher-guided traces~\citep{chen2025retaining}.

\textbf{Models and generation.}
We use two distilled reasoning models as students: DeepSeek-R1-Distill-Qwen-7B and DeepSeek-R1-Distill-LLaMA-3-8B. For each question in a shared set of 1,000 open-ended reasoning problems, we generate outputs with temperature 0.7 and sample multiple random seeds to obtain diverse reasoning traces. For teacher-guided baselines, we use DeepSeek-R1-0528 only as the trace generator.

\textbf{Best-of-$K$ filtering.}
We apply a Best-of-$K$ ($K=4$) filtering strategy to construct successful on-policy traces. We keep only candidates whose final answers match the ground truth under task-specific normalization. Among correct traces, we select \emph{intermediate-length} samples rather than always taking the shortest or longest outputs. This avoids trivial short traces and excessively verbose outliers, yielding a more balanced supervision set.

\textbf{Why self-distilled.}
The goal of self-distilled trace construction is to keep fine-tuning on-policy. In low-data LoRA fine-tuning, teacher-generated traces may be stylistically and behaviorally out-of-distribution for the student model. By contrast, self-distilled traces are drawn from the student's own reasoning distribution, which helps reduce distributional shift and preserve its native reasoning behavior during adaptation.

\subsection{Structured Reasoning Interface}

The second component of \frameworkabbr\ is a structured interface over each reasoning trace. This interface is built in three stages: node segmentation, taxonomy annotation, and reasoning-tree construction. Together, these steps produce a shared representation that supports both ECN pruning and before-or-after analysis of reasoning traces.

\textbf{Node segmentation.}
We segment each reasoning trace into a node sequence using a deterministic, rule-based procedure. The goal is to create units: (i) large enough to preserve local semantic coherence, but (ii) small enough to support annotation, pruning, and analysis.

\textbf{Segmentation rules.}
We detect discourse markers such as \textit{However, But, Alternatively, So, Now} (case-insensitive) and punctuation-based boundaries such as sentence endings and line breaks. When a marker appears at the start of a sentence, we split the trace so that the full marker-starting sentence becomes a new node, while the preceding text remains in the previous node. Because pruning is defined only at node boundaries, this segmentation helps preserve semantic continuity and avoids producing fractured supervision targets.

\textbf{Taxonomy annotation.}
Each node is annotated with a functional role that captures how it contributes to the reasoning process. We adopt the following 5-class taxonomy:
\begin{itemize}[noitemsep, topsep=0pt, leftmargin=10pt]
    \item \textbf{Backtracking:} revising earlier steps or assumptions to correct errors or resolve conflicts.
    \item \textbf{Verification:} checking or validating a claim or result without changing the main plan.
    \item \textbf{Exploration:} proposing or trying alternative hypotheses or solution routes.
    \item \textbf{Clarification:} restating assumptions, reducing ambiguity, or organizing problem.
    \item \textbf{Conclusion:} asserting an intermediate or final result derived from prior reasoning.
\end{itemize}

% We conduct two independent taxonomy-annotation experiments: uses two different annotation models \texttt{Gemini-2.5-Flash} and \texttt{DeepSeek-V4-Flash}. Both use the same few-shot prompt, input traces, label schema, and deterministic decoding. The model outputs a label and brief justification for audit.

\textbf{Reasoning tree construction.}
On top of the node sequence, we construct an n-ary reasoning tree that provides a global structural view of the trace. Each tree node corresponds to one segmented reasoning unit, and a parent--child edge indicates that one node supports another as evidence, premise, sub-result, or elaboration. Although \ECN\ is defined on the ordered node sequence, the tree is useful for analyzing structural redundancy and comparing reasoning behavior before and after pruning.

Taxonomy annotation and tree construction are performed automatically using powerful annotation models. We adopt two independent annotation models: \texttt{DeepSeek-V4-Flash} and \texttt{Gemini-2.5-Flash}. Both use the same few-shot prompt, input traces, label schema, and deterministic decoding. The annotation model outputs a label and brief justification for audit. These two annotation models lead to the consistent findings. Due to space constraints, results in the main table are from \texttt{Gemini-2.5-Flash} and results from \texttt{DeepSeek-V4-Flash} are introduced in the Appendix~\ref{app:additional results}.

\subsection{Earliest Correct Node Pruning Strategy}

\textbf{Earliest Correct Node (ECN) identification.}
Built on top of the structured node sequence, \ECN\ requires identifying the earliest node that is both (i) an answering conclusion and (ii) correct. Among nodes labeled Conclusion, we further distinguish between intermediate conclusions, which support later reasoning but do not directly answer the question $q$, and answering conclusions, which are intended to answer $q$, even if the model later revises them. Given the ground-truth answer $a^\star$, we judge whether an answering-conclusion node entails the correct final answer. When the node contains an explicit answer, we extract the value and compare it with $a^\star$ under task-specific normalization. When the node is implicit or ambiguous, we prompt annotation model with $(q, a^\star, N_i)$ to determine whether $N_i$ constitutes a correct answering conclusion. We further validate the results with human annotation in the appendix.

% \textbf{ECN definition.}
% Let $\operatorname{CAC}(N_i)$ indicate that node $N_i$ is judged to be a correct answering conclusion. We define
% \[
% \mathcal{C} =
% \{\, i \in \{1,\dots,m\} : \operatorname{CAC}(N_i) \,\}.
% \]
% If $\mathcal{C} \neq \varnothing$, the \ECN\ index is defined as
% \[
% k = \min \mathcal{C}.
% \]
% That is, \ECN\ selects the earliest node in the trace that already yields the correct answer. If $\mathcal{C} = \varnothing$, the trace is excluded from ECN supervision.

% \textbf{ECN pruning.}
% Given the node sequence $S=(N_1,\dots,N_m)$ and the \ECN\ index $k$, ECN pruning keeps the minimal prefix $\text{ECN}(S)=(N_1,\dots,N_k)$. Because pruning operates on coherent reasoning nodes rather than arbitrary token spans, the resulting target remains semantically continuous while removing redundant post-solution reasoning. 

\begin{table*}[t]
\centering
\small
\setlength{\tabcolsep}{2.5pt}
\begin{tabular}{ll|cc cc|cc cc|cc cc}
\toprule
& & \multicolumn{4}{c|}{\textbf{AIME'24}} & \multicolumn{4}{c|}{\textbf{GSM8K}} & \multicolumn{4}{c}{\textbf{Math 500}} \\
\cmidrule(lr){3-6} \cmidrule(lr){7-10} \cmidrule(lr){11-14}
\textbf{Model} & \textbf{Method}
& \textbf{Acc} & \textbf{$\Delta$} & \textbf{Tok} & \textbf{$\Delta$\%}
& \textbf{Acc} & \textbf{$\Delta$} & \textbf{Tok} & \textbf{$\Delta$\%}
& \textbf{Acc} & \textbf{$\Delta$} & \textbf{Tok} & \textbf{$\Delta$\%} \\
\midrule
\multirow{9}{*}{Qwen}
& Base      & .367 & --     & 10848 & --      & .901 & --     & 1311 & --      & .878 & --     & 3678 & -- \\
& NT        & .233 & -.134  & 2411  & -77.8   & .873 & -.028  & 192  & -85.4   & .772 & -.106  & 678  & -81.6 \\
& TG-Full   & .067 & -.300  & 14854 & +36.9   & .535 & -.366  & 2548 & +94.4   & .278 & -.600  & 6436 & +75.0 \\
& TG-RP     & .033 & -.334  & 15344 & +41.5   & .427 & -.474  & 6153 & +369.4  & .186 & -.692  & 7319 & +99.0 \\
& TG-ECN    & .100 & -.267  & 13574 & +25.1   & .792 & -.109  & 1168 & -10.9   & .540 & -.338  & 4512 & +22.7 \\
& SD-Full   & \textbf{.533} & \textbf{+.166}  & 9784  & -9.8    & \textbf{.912} & \textbf{+.011}  & 1174 & -10.5   & \textbf{.876} & \textbf{-.002}  & 2989 & -18.7 \\
& SD-Token Skip & .133 & -.234  & 13353 & +23.1   & .645 & -.256  & 3645 & +178.0  & .342 & -.536  & 6252 & +70.0 \\
& SD-RP     & .233 & -.134  & 8069 & -25.6 & .884 & -.017 & 1003 & -23.5 & .818 & -.060 & \textbf{1992} & \textbf{-45.8} \\
\rowcolor{gray!15}
& SD-ECN (ours) & .467 & +.100  & \textbf{7716} & \textbf{-28.9} & .911 & +.010 & \textbf{946} & \textbf{-27.8} & .868 & -.010 & 2118 & -42.4 \\
\midrule
\multirow{9}{*}{Llama}
& Base      & .433 & --     & 11855 & --      & .840 & --     & 1535 & --      & .879 & --     & 3413 & -- \\
& NT        & .033 & -.400  & 3988  & -66.4   & .764 & -.076  & 257  & -83.3   & .608 & -.271  & 989  & -71.0 \\
& TG-Full   & .067 & -.366  & 15332 & +29.3   & .613 & -.227  & 4138 & +169.6  & .436 & -.443  & 5710 & +67.3 \\
& TG-RP     & .033 & -.400  & 16167 & +36.4   & .606 & -.234  & 4617 & +200.8  & .310 & -.569  & 6650 & +94.9 \\
& TG-ECN    & .167 & -.266  & 14353 & +21.1   & .667 & -.173  & 3292 & +114.5   & .430 & -.449  & 5487 & +60.8 \\
& SD-Full   & \textbf{.600} & \textbf{+.167}  & 9948  & -16.1   & .822 & -.018  & 1514 & -1.4    & .822 & -.057  & 3322 & -2.7 \\
& SD-Token Skip & .033 & -.400 & 15735 & +32.7   & .674 & -.166  & 3248 & +111.6  & .496 & -.383  & 5418 & +58.8 \\
& SD-RP     & .233 & -.200  & \textbf{5170} & \textbf{-56.4} & .807 & -.033 & \textbf{784} & \textbf{-48.9} & .740 & -.139 & \textbf{1628} & \textbf{-52.3} \\
\rowcolor{gray!15}
& SD-ECN (ours) & .400 & -.033  & 9555 & -19.4 & \textbf{.832} & \textbf{-.008} & 920 & -40.1 & \textbf{.877} & \textbf{-.002} & 2168 & -36.5 \\
\bottomrule
\end{tabular}
\caption{Main results showing the trade-off between reasoning accuracy and token efficiency across models, benchmarks, and training/pruning strategies. Method abbreviations are: NT = No Thinking; TG = Teacher-Guided; SD = Self-distilled; RP = Random Prune. Our main method, \textbf{SD-ECN}, is the primary instantiation of \frameworkabbr. $\Delta$ denotes the accuracy change relative to the corresponding Base model, and $\Delta$\% denotes the relative token change. Best self-distilled results are shown in \textbf{bold}.}
\label{tab:main_results}
\end{table*}
% \vspace{-2mm}
\section{Experiment Setup}
% \vspace{-2mm}
\textbf{Models.}
We use two distilled reasoning models as students: DeepSeek-R1-Distill-Qwen-7B and DeepSeek-R1-Distill-LLaMA-3-8B. For self-distilled variants, the same student model is used both to generate training traces and to initialize subsequent LoRA fine-tuning. Teacher-guided baselines use DeepSeek-R1-0528 to generate the trace. This design isolates the effect of supervision source and pruning strategy from changes in model family or initialization.

% \vspace{-0.35em}
% \paragraph{Data.}
\textbf{Training data.} We randomly sample 1000 questions from PRM-12 and refer to this setting as \emph{low-data} fine-tuning. For each question, we generate candidate reasoning traces, retain successful traces using Best-of-$K$ filtering, and transform the retained traces with \frameworkabbr: deterministic node segmentation, taxonomy annotation, reasoning-tree construction, and ECN pruning. The resulting traces are used as SFT targets.

\textbf{Evaluation benchmarks.} We evaluate both problem-solving performance and reasoning efficiency on three benchmarks with different difficulty levels and reasoning styles: GSM8K, Math~500, and AIME~2024. Following prior work~\citep{chen2024not}, we evaluate the performance based on the following metrics. 1. Answer Accuracy.
For open-ended questions, correctness is determined by matching the final numerical answer with the ground truth. 2. Token quantity. We report the average number of tokens used in the complete output.

% \vspace{-0.35em}
\textbf{Structured trace interface and annotation.}
Our training targets are produced through the shared structured reasoning interface introduced in Section~\ref{sec:method}. We first segment each trace \emph{deterministically} into semantically coherent reasoning nodes using discourse and formatting rules. We then use annotation model to assign taxonomy labels, construct reasoning trees, and, when needed, judge ambiguous candidate conclusion nodes for ECN identification. Thus, the same interface is used both to construct ECN-pruned supervision targets and to analyze reasoning behavior before and after pruning. Appendix~A provides all prompts, decoding settings, structured output formats, and post-processing rules.

% \vspace{-0.35em}
\textbf{Implementation details}
We apply LoRA-based~\citep{hu2022lora} supervised fine-tuning on a single NVIDIA A100 (80GB) GPU with a maximum training sequence length of 8192 tokens. Unless otherwise specified, all hyperparameters are fixed across models and datasets; Full details are given in Appendix~\ref{app:reproducibility}. 

% \paragraph{Reproducibility.}
% BECNuse the quality of the training targets depends on the structured trace-processing pipeline, reproducibility is especially important in our setting. We therefore provide the full annotation prompts, intermediate examples, post-processing scripts, and training hyperparameters in the appendix and supplementary materials, and we will release the code for data construction, pruning, and fine-tuning.

\textbf{Baselines.} We compare the following baselines:
\begin{itemize}[noitemsep, topsep=1pt, leftmargin=10pt]
    \item Base: the original student model without additional fine-tuning.
    \item No Thinking: fine-tuning on outputs with all content inside the \texttt{<think>} tag removed, retaining only the final answer format.
    \item Teacher-Guided-Full: fine-tuning on full reasoning traces generated by the teacher model.
    \item Teacher-Guided Random Prune: fine-tuning on teacher-guided traces truncated at a random node, controlling for length reduction without a principled stopping rule.
    \item Teacher-Guided-ECN: fine-tuning on teacher-generated traces after ECN pruning.
    \item Self-distilled-Full: fine-tuning on full reasoning traces generated by the student itself.
    \item Self-distilled Token Skip~\citep{xia2025tokenskip}: a compression baseline that shortens self-distilled traces through token-skipping style CoT compression.
    \item Self-distilled Random Prune: fine-tuning on self-distilled traces truncated at a random node boundary, controlling for length reduction without a principled stopping rule.
    \item Self-distilled-ECN (ours): fine-tuning on self-distilled traces after ECN pruning.
\end{itemize}

% \vspace{-0.8em}
\section{Results}

\paragraph{\frameworkabbr\ achieves strong efficiency--accuracy trade-offs across models and task difficulty.}
Table~\ref{tab:main_results} compares our main method, Self-distilled-ECN, against the base models and alternative supervision/pruning strategies on GSM8K, Math~500, and AIME~2024, spanning easy, medium, and hard reasoning settings. On GSM8K, Self-distilled-ECN improves accuracy while reducing token usage for both models. For Qwen, it reaches 91.1\% accuracy versus 90.1\% for Base while reducing tokens by 27.8\%. For Llama, it attains 83.2\% accuracy versus 84.0\% for Base while reducing tokens by 40.1\%. On Math~500, Qwen reaches 86.8\% accuracy versus 87.8\% for Base with a 42.4\% token reduction. For Llama, it reduces tokens by 36.5\%, while accuracy decreases from 87.9\% to 84.4\%. On AIME~2024, it improves Qwen accuracy from 36.7\% to 46.7\% while reducing tokens by 28.9\%; for Llama, it reduces tokens by 19.4\%, though with a modest accuracy drop from 43.3\% to 40.0\%. Overall, these results show that STOP consistently mitigates overthinking, while the efficiency-accuracy trade-off becomes more pronounced on harder problems.

% \vspace{-0.8em}
\paragraph{ECN is consistently stronger than alternative pruning and compression baselines.}
Across both model families, ECN is the most effective pruning rule among the tested length-reduction baselines. Compared with Random Pruning and Token Skip, ECN achieves dramatically higher accuracy while using fewer tokens on nearly all benchmarks. For example, on Math~500 with Qwen, Self-distilled-ECN attains 86.8\% accuracy with 2118 tokens, compared with 34.2\% and 6252 tokens for Token Skip, and 81.8\% and 1992 tokens for Random Pruning. A similar pattern holds for LLaMA, where Self-distilled-ECN reaches 87.7\% accuracy with 2168 tokens, far outperforming both Token Skip (49.6\%, 5418 tokens) and Random Pruning (74.0\%, 1628 tokens). These comparisons show that the benefit of ECN does not come from shortening traces arbitrarily; it comes from stopping at a coherent node-level conclusion.

ECN is also effective beyond the self-distilled setting. When applied to teacher-guided traces, Teacher-Guided-ECN consistently reduces token usage relative to Teacher-Guided-Full and often improves accuracy as well. For Qwen on GSM8K, Teacher-Guided-ECN improves accuracy from 53.5\% to 79.2\% while reducing tokens from 2548 to 1168. On Qwen Math~500, it improves accuracy from 27.8\% to 54.0\% while reducing tokens from 6436 to 4512. Similar gains appear on AIME~2024 for both Qwen and LLaMA. This shows that ECN is a robust pruning strategy across multiple supervision sources, even though its strongest performance is obtained in the full self-distilled STOP framework.

\begin{figure}[!tbp]
    \centering
    \includegraphics[width=0.90\linewidth]{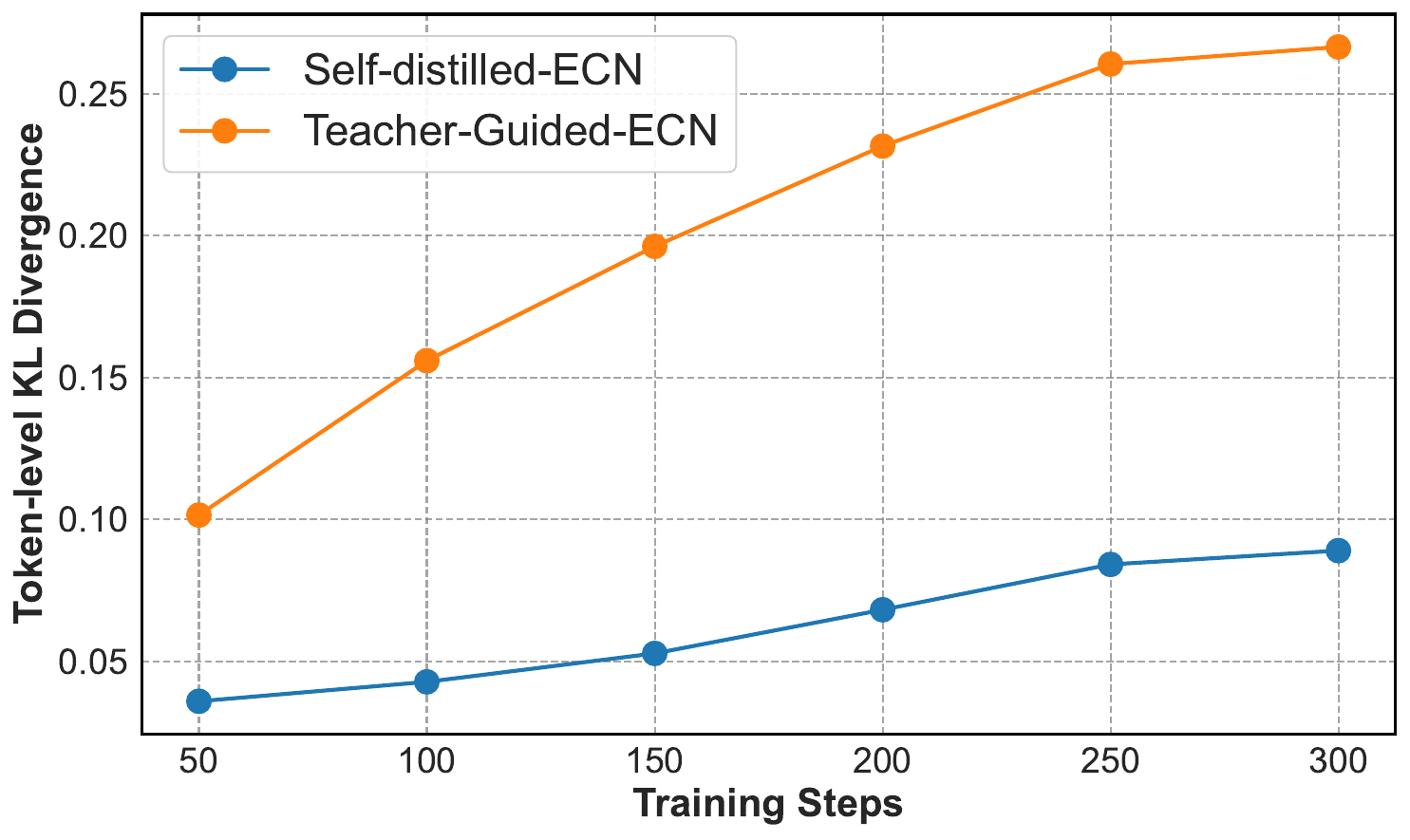}
    \caption{Token-level forward KL divergence between the reference model (DeepSeek-R1-Distill-Qwen-7B) and models fine-tuned with Self-distilled-ECN and Teacher-Guided-ECN.}
    \label{fig:kl_checkpoint}
\end{figure}

% \begin{figure}[t]
%     \centering
%     \includegraphics[width=\linewidth]{fig/qwen_low_data_deltas_paper.pdf}
%  % fig/self_vs_teacher_low_data_deltas_paper.pdf}
%     \caption{Accuracy gains and token savings of self-distilled models across three supervision budgets (200, 500, and Full) on AIME 2024, GSM8K, and MATH-500 for Qwen. $\Delta$Acc. = self-distilled $-$ teacher-guided, and $\Delta$Tok. = teacher-guided $-$ self-distilled, where larger positive $\Delta$Tok. indicates greater token savings.}
%     \label{fig:self-distilled-low-data-deltas}
% \end{figure}

% \vspace{-0.8em}
\paragraph{Self-distilled supervision is substantially more effective than teacher-guided supervision in the low-data regime.}
A pattern in Table~\ref{tab:main_results} is that self-distilled supervision outperforms teacher-guided supervision under both full-trace and ECN training. For Qwen on AIME~2024, Self-distilled-Full reaches 46.7\% accuracy compared to 6.7\% for Teacher-Guided-Full, and Self-distilled-ECN reaches 50.0\% compared to 10.0\% for Teacher-Guided-ECN. The same pattern holds across GSM8K and Math~500 for both student models: self-distilled variants are consistently stronger than their teacher-guided counterparts, often by large margins. This indicates that in the low-data setting, staying close to the student's native reasoning distribution is more important than imitating a stronger but mismatched teacher.

To examine this effect at the token-distribution level, Figure~\ref{fig:kl_checkpoint} measures the forward KL divergence between the base model and models fine-tuned with self-distilled ECN or teacher-guided ECN. The teacher-guided model undergoes a larger distributional shift, with the KL divergence increasing from around 0.1 at 50 training steps to above 0.25 by 300 steps. In contrast, the self-distilled model remains closer to the base model throughout training, staying roughly in the 0--0.1 range. This suggests that self-distilled ECN preserves the original reasoning behavior of the student model better than teacher-guided supervision, helping explain why self-distilled STOP is more stable and effective in low-data fine-tuning.

\begin{figure}[!tbp]
    \centering
    \includegraphics[width=\linewidth]{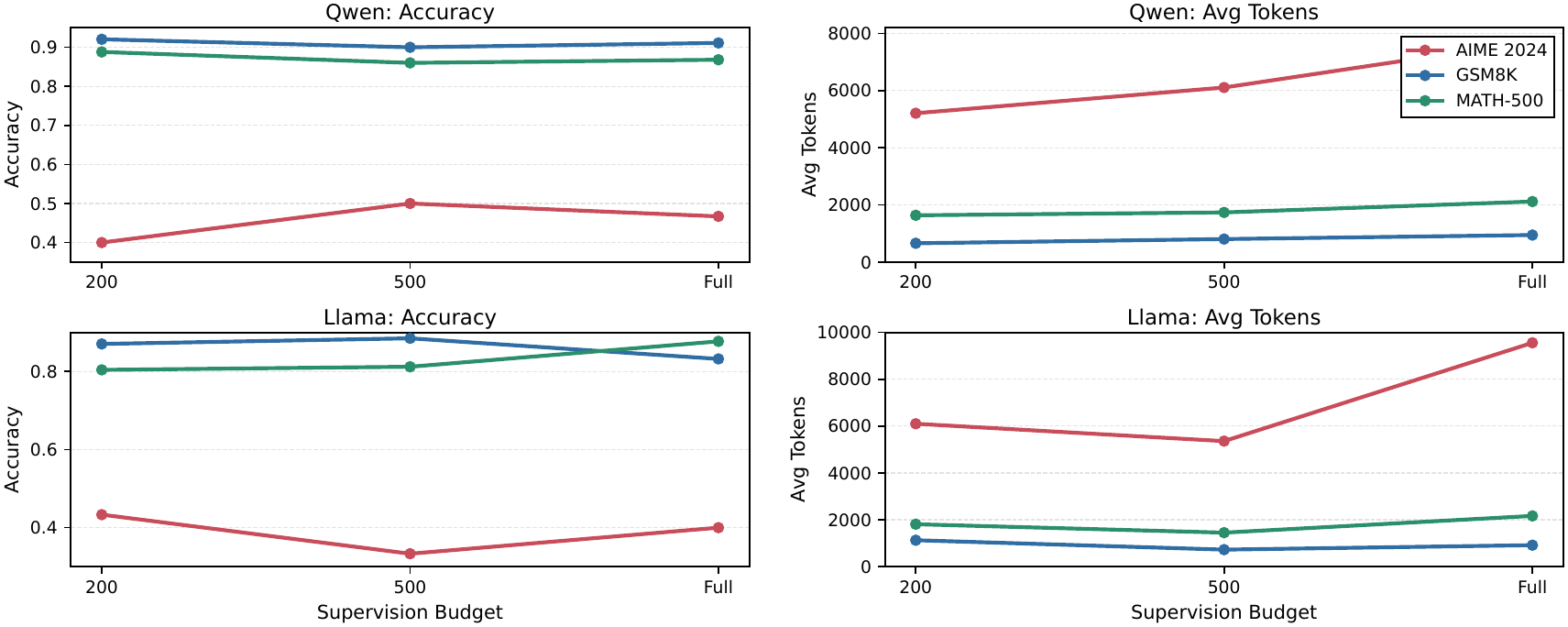}
 % fig/self_vs_teacher_low_data_deltas_paper.pdf}
    \caption{Accuracy gains and token savings of self-distilled models across three supervision budgets (200, 500, and Full) on AIME 2024, GSM8K, and MATH-500 for Qwen and Llama.}
    \label{fig:self-distilled-low-data-deltas}
\end{figure}

To examine how self-distilled supervision behaves under different supervision budgets, Figure~\ref{fig:self-distilled-low-data-deltas} plots performance at 200, 500, and the full dataset (about 900 examples). The main pattern is that scaling the budget does not produce a uniform monotonic gain across models and benchmarks. For Qwen, the 200-example setting is already highly competitive, matching or exceeding larger budgets on GSM8K and MATH-500, while AIME 2024 peaks at 500 examples. For Llama, the 500-example setting gives the best balance on GSM8K, whereas the full budget is most beneficial on MATH-500. Token usage generally increases with larger supervision budgets, although some mid-budget settings remain more efficient than the full setting. Overall, these results suggest that self-distilled supervision is already effective in the low-data regime, and that increasing the budget beyond a few hundred examples does not consistently translate into better efficiency or accuracy.

% \begin{figure}[t]
%     \centering
%     \includegraphics[width=\linewidth]{fig/self_vs_teacher_low_data_deltas_paper.pdf}
% \caption{
% Accuracy gains and token savings of self-distilled models relative to teacher-guided baselines across three supervision budgets (200, 500, and Full) on AIME 2024, GSM8K, and MATH-500 for Qwen and Llama. $\Delta$Acc. = self-distilled $-$ teacher-guided, and $\Delta$Tok. = teacher-guided $-$ self-distilled, where larger positive $\Delta$Tok. indicates greater token savings.
% }
%     \label{fig:self-distilled-low-data-deltas}
% \end{figure}

% \vspace{-0.8em}
% \paragraph{Coherent stopping matters more than raw length reduction.}
% The sanity-check baselines further show that reducing output length alone is not sufficient. Random Pruning severely harms both accuracy and efficiency: for example, on Math~500 with LLaMA, Random Pruning drops accuracy from 87.9\% to 31.0\% and increases average generation from 3413 to 6650 tokens. No-thinking minimizes token usage, but does so by removing reasoning entirely and therefore incurs substantial accuracy drops across all benchmarks. Together, these baselines reinforce the central design principle of \frameworkabbr: the goal is not merely to shorten traces, but to stop at a semantically coherent point that preserves useful reasoning while removing redundant post-solution computation.

% \vspace{-2mm}

\begin{figure}[!tbp]
    \centering
    \includegraphics[width=0.84\linewidth]{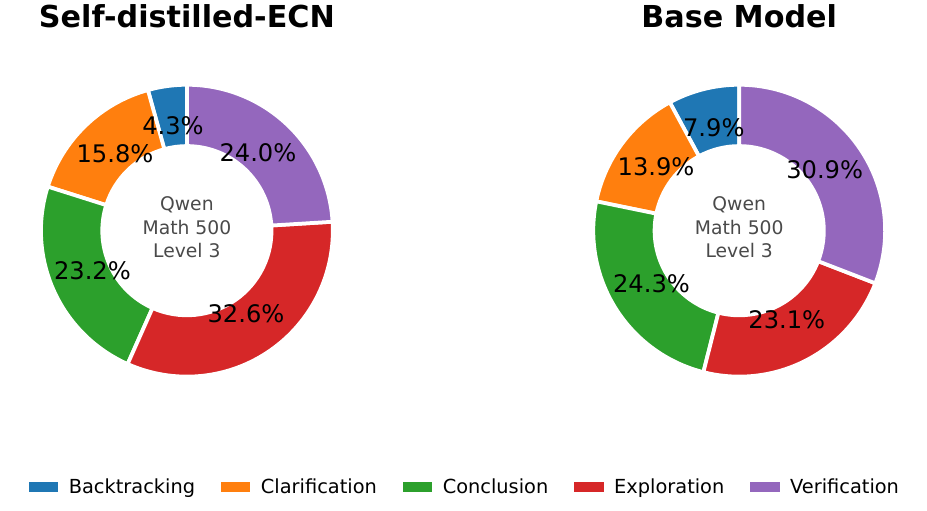}
    \caption{Reasoning node-type distribution on Math~500 Level~3 for Qwen Base and Self-distilled-ECN. Self-distilled-ECN shifts reasoning toward more \textit{Exploration} and less \textit{Verification}/\textit{Backtracking}, while largely preserving \textit{Conclusion}.}
    \label{fig:reasoning_node-distribution}
\end{figure}

\section{Analysis}

Rather than judging \frameworkabbr\ solely by token reduction, we examine whether it changes the \emph{structure} of reasoning~\citep{besta2025demystifying}. Using three views of self-distilled \texttt{Gemini-2.5-Flash} traces—taxonomic motifs~\citep{gandhi2025cognitive}, morphological patterns~\citep{jiang2025makes}, and pruning efficiency dynamics~\citep{Wei2025TheEO}—we find that \frameworkabbr\ does more than compress outputs: it reorganizes reasoning into a more focused, stable, and structurally efficient form.

\textbf{Taxonomic motifs: \frameworkabbr\ reallocates reasoning from redundant checking to productive search.}
Figure~\ref{fig:reasoning_node-distribution} compares the node-type distribution of the Base Qwen model and the Self-distilled-ECN Qwen model on Math~500 Level~3. Self-distilled-ECN substantially increases \textit{Exploration} from 23.1\% to 32.6\% (+9.5 points), suggesting that the model devotes more of its reasoning budget to core solution search rather than post-hoc deliberation. At the same time, \textit{Verification} decreases from 30.9\% to 24.0\% (-6.9 points) and \textit{Backtracking} drops from 7.9\% to 4.3\% (-3.6 points), consistent with ECN removing late-stage re-checking and unnecessary revisions after a correct answering conclusion has already appeared. Importantly, \textit{Conclusion} remains nearly unchanged (24.3\% to 23.2\%), and \textit{Clarification} slightly increases (13.9\% to 15.8\%), indicating that STOP preserves the core logical scaffolding while changing the semantic flow of reasoning. Thus, the gain is not mere compression; it is a redistribution of reasoning effort toward higher-yield cognitive operations.

\textbf{Morphological patterns: \frameworkabbr\ transforms ``bushy'' exploration into streamlined, goal-directed reasoning trees.} Table~\ref{tab:tree-structure-main} shows that \frameworkabbr\ shifts the Base model from broad, hesitant branching to a more linear and efficient search strategy. Trees become more compact (15.20 $\rightarrow$ 11.11 nodes) mainly by reducing lateral expansion: maximum width drops sharply (5.23 $\rightarrow$ 3.47) while depth remains similar (6.28 $\rightarrow$ 5.62). As a result, the width-to-depth ratio falls from 1.15 to 0.93, indicating a move from parallel hypothesis testing toward more confident deduction, reinforced by lower mean and maximum branching factors. The trees are also more balanced and efficient, with structural efficiency rising from 0.53 to 0.65 and Sackin and Generalized Colless indices falling substantially (30.91 $\rightarrow$ 16.73; 11.29 $\rightarrow$ 5.11). Overall, \frameworkabbr\ not only shortens reasoning traces but also concentrates computation on the essential inferential path.

\begin{table}[!tbp]
\centering
\scriptsize
\setlength{\tabcolsep}{3pt}
\renewcommand{\arraystretch}{0.95}
\begin{tabular}{@{}p{0.52\linewidth}ccc@{}}
\toprule
Metric & ECN & Base & $\Delta$ \\
\midrule
Total tree nodes & 11.11 & 15.20 & -4.10 \\
Maximum depth & 5.62 & 6.28 & -0.66 \\
Maximum width & 3.47 & 5.23 & -1.76 \\
Width-to-depth ratio & 0.93 & 1.15 & -0.22 \\
Mean leaf depth & 4.29 & 4.60 & -0.31 \\
Leaf depth variance & 4.23 & 2.59 & 1.65 \\
Mean branching factor & 1.69 & 1.90 & -0.21 \\
Maximum branching factor & 3.23 & 4.61 & -1.38 \\
Branching variance & 1.81 & 3.43 & -1.62 \\
Structural efficiency & 0.65 & 0.53 & 0.13 \\
Sackin index & 16.73 & 30.91 & -14.17 \\
Generalized Colless index & 5.11 & 11.29 & -6.17 \\
\bottomrule
\end{tabular}
\caption{Morphology metrics on Math~500 Level~3. See details in Appendix ~\ref{sec:metric_definitions}.}
\label{tab:tree-structure-main}
\end{table}

\begin{figure}[!tbp]
    \centering
    \includegraphics[width=\linewidth]{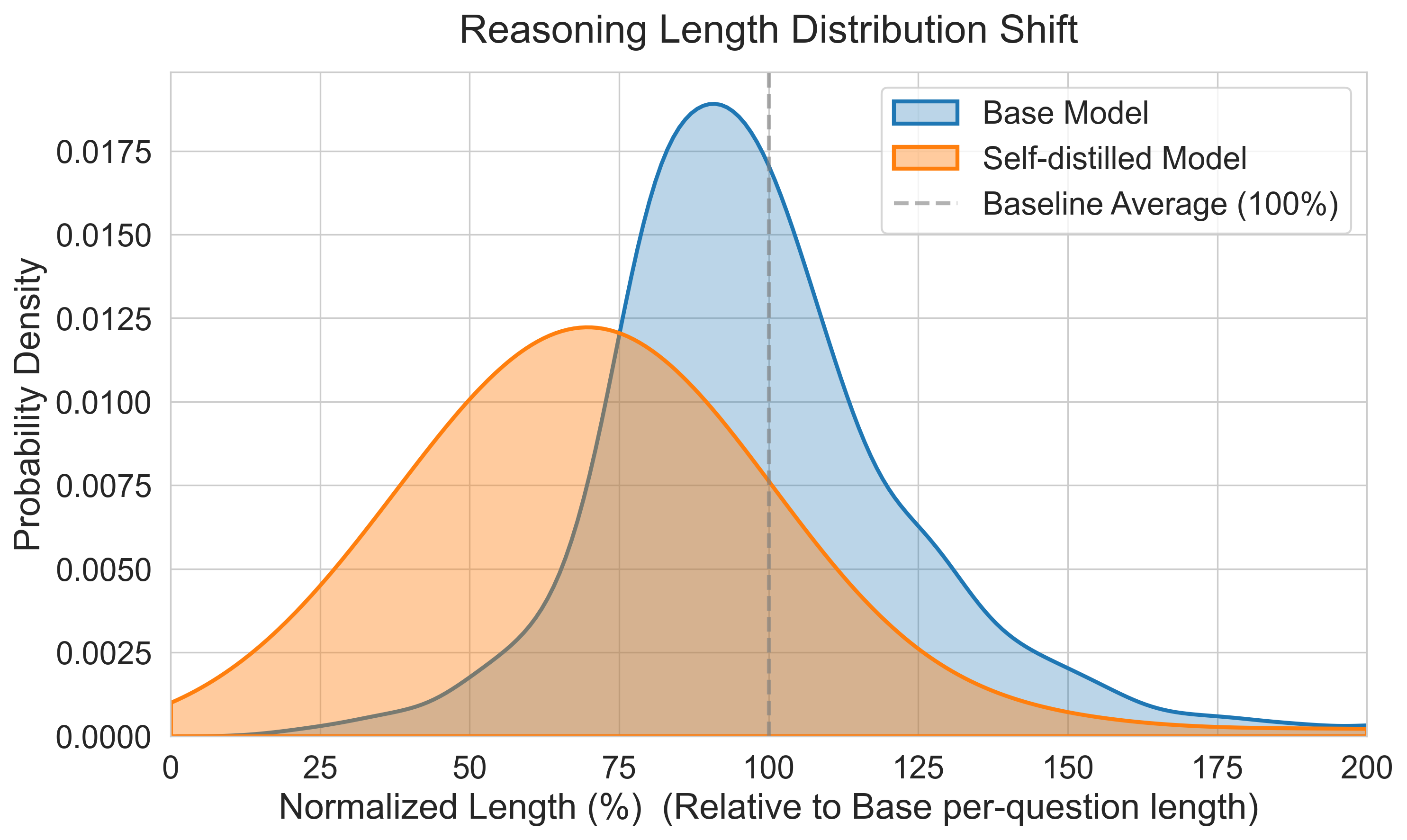}
    \caption{Response-length distribution of the base Qwen model and the self-distilled ECN-Qwen model on Math~500 (Level~3).}
    \label{fig:math500_token_scatter}
\end{figure}

% \vspace{-0.4em}
\textbf{Pruning efficiency dynamics: \frameworkabbr\ shortens traces through stable structural adaptation rather than brittle truncation.} Appendix Table~\ref{tab:search-pruning} shows that the Earliest Final-Answer depth changes little from Base to Self-distilled-ECN (3.029 $\rightarrow$ 3.181), indicating that \frameworkabbr\ does not reduce cost mainly by making answers appear much earlier. Instead, it cuts redundant reasoning after the answer first emerges: the post-answer source-node gap drops sharply from 12.143 to 5.762. This pattern aligns with Figure~\ref{fig:math500_token_scatter}, where normalized reasoning-length distributions shift left under Self-distilled-ECN, showing a broad move toward shorter outputs rather than a few extreme truncations. Together, these results suggest that \frameworkabbr\ changes the model’s default reasoning policy by preserving similar answer-emergence depth while substantially reducing unnecessary post-answer expansion.

% \vspace{-2mm}
\section{Related Work}

% \vspace{-0.8em}
\textbf{Overthinking and Reasoning-Length Control} Recent reasoning language models (RLMs) achieve strong performance on multi-step tasks by generating long chain-of-thought (CoT) traces \citep{jaech2024openai, guo2025deepseek, 10.1609/aaai.v37i2.25264}. However, extended reasoning often introduces substantial computational overhead and frequently contains repetitive or unnecessary steps, a phenomenon widely referred to as \emph{overthinking} \citep{sui2025stop}. Prior studies analyze this behavior and its failure modes across modern long-reasoning systems \citep{team2025kimi, chen2024not}. A natural direction is to \emph{compress} verbose natural-language CoT into shorter representations while preserving problem-solving ability. For example, \citet{shen2025codi} propose continuous chain-of-thought via self-distillation, aiming to move from explicit natural-language reasoning to a more compact internal representation.  Another line of work reduces reasoning cost by \emph{pruning} particular parts of the reasoning trace or discouraging specific behaviors. \citet{wang2025thoughts} observe that o1-style models may jump between reasoning paths instead of fully developing promising ones, suggesting that a large fraction of tokens can be spent on low-yield branches. Building on this perspective, \citet{luo2025o1} propose O1-Pruner to reduce reasoning cost.

\textbf{Teacher-Guided vs. Self-Distilled Supervision.}
A common strategy for reasoning adaptation is teacher-guided distillation, where a stronger teacher model provides high-quality reasoning traces as supervision for a student model \citep{li2025naturalthoughts}. While effective, this setup often introduces distribution mismatch, since teacher and student models may differ substantially in capability, style, verbosity, and intermediate reasoning preferences, which can lead to unstable optimization or behavioral drift under direct trace imitation \citep{li2025small}. In contrast, self-distillation and on-policy supervision use trajectories that are more consistent with the student’s own distribution, thereby mitigating such mismatch and providing a more stable supervision signal \citep{chen2025retaining,agarwal2024policy}. This distinction is especially relevant in the low-data fine-tuning regime, where limited supervision leaves less room to accommodate teacher-student mismatch. Our method follows this intuition by constructing pruning targets from the student’s own reasoning traces rather than externally provided teacher trajectories.

% \vspace{-0.8em}
\section{Conclusion}
% \vspace{-0.5em}

We introduced \frameworkabbr, a unified on-policy framework for improving the efficiency of long-form reasoning in low-data regimes. STOP combines self-distilled trace construction, a structured reasoning interface based on node segmentation, taxonomy annotation, and tree construction, and ECN pruning, a node-level stopping rule that removes redundant post-solution reasoning while preserving semantic continuity. Across two distilled reasoning models and three reasoning benchmarks, STOP reduces generated tokens by 19.9--41.9\% while maintaining competitive accuracy. Beyond efficiency, our analyses show that STOP systematically reorganizes the model's reasoning behavior: it produces more focused reasoning structures, reallocates effort away from redundant verification and backtracking, and preserves greater on-policy stability than teacher-guided supervision. These results suggest that efficient reasoning is not just a matter of cutting tokens, but of learning when and how to stop thinking productively.

\section*{Limitations}

\paragraph{ECN is an oracle training-time stopping rule.}
ECN is defined using answer correctness: identifying the earliest \emph{correct} answering conclusion requires access to the ground-truth answer or an external correctness judge. This makes ECN a strong supervision-construction method, but not a directly deployable inference-time stopping rule. A practical deployment version would require confidence estimation, learned stopping predictors, or verifier-based approximations.

\paragraph{The structured interface depends partly on external annotation.}
While node segmentation is deterministic, taxonomy annotation, tree construction, and some ambiguous conclusion judgments rely on annotation models. This introduces dependence on an external model with its own biases, errors, and cost. Mistakes in labeling or conclusion judgment can propagate into pruning decisions and affect the quality of the resulting supervision targets.

\bibliography{custom}

\appendix

\lstset{
  basicstyle=\ttfamily\small,
  breaklines=true,
  columns=fullflexible,
  frame=single,
  showstringspaces=false,
  aboveskip=0.5em,
  belowskip=0.5em
}

\newpage

\appendix

\section{Reproducibility Details}
\label{app:reproducibility}

\subsection{Training Details}

We fine-tune \texttt{DeepSeek-R1-Distill-Qwen-7B} and  \texttt{DeepSeek-R1-Distill-Llama-8B} using LoRA with \texttt{bfloat16} precision. The maximum sequence length is 8192, the random seed is 1, and the per-device train/eval batch size is 1/1 with \texttt{gradient\_accumulation\_steps=16}. We use a learning rate of \(1\times10^{-4}\), LoRA rank 8, LoRA alpha 32, and a warmup ratio of 0.05. Evaluation, checkpoint saving, and logging are performed every 100, 100, and 5 steps, respectively.

For the experiments reported in Table~\ref{tab:main_results}, we train self-distilled models for 6 epochs but teacher-guided models for 2 epochs. This choice is intentional: teacher-guided training showed severe overfitting at 6 epochs, and using 2 epochs yields a stronger and fairer teacher-guided baseline. However, for the supervision-budget comparison in Figure~\ref{fig:self-distilled-low-data-deltas}, teacher-guided models are trained for 6 epochs to keep the training schedule consistent across budget settings.

\subsection{Inference Details}

At inference time, we set the maximum generation length to 16384 tokens for AIME~2024 and 8192 tokens for GSM8K and Math~500 to avoid artificial truncation. We use the same decoding configuration across compared methods for fair evaluation. 

\subsection{Annotation Model Configuration}
\label{app:gemini}

Our annotation pipeline is tested on two language models,
\textbf{Gemini-2.5-Flash} and \textbf{Deepseek-V4-Flash}, which are used to perform the structured
annotation tasks in our pipeline, including taxonomy labeling,
reasoning tree construction, and ECN conclusion correctness
judgment.

To improve reproducibility, we use a fixed decoding configuration
across all annotation stages.

\textbf{Model.} Gemini-2.5-Flash (Google Gemini API); Deepseek-V4-Flash (Deepseek API).

\textbf{Decoding configuration.} Temperature is set to 0.0, while
top-$p$ and top-$k$ sampling are disabled. We do not impose a strict
limit on the maximum output tokens, and the thinking budget is left
unrestricted during annotation.

\textbf{API usage and cost.} All annotation calls are executed through
the official Google Gemini Batch API or DeepSeek Official API. The same model version and
decoding configuration are used throughout the entire dataset
construction pipeline. The total API cost for processing the 1000
training examples is less than \$20.

\subsection{Annotation Prompts}
\label{app:prompts}

In this section, we provide the prompts used in our model-assisted annotation pipeline, including taxonomy labeling, tree construction, and ECN conclusion judgment.

\textbf{Taxonomy labeling prompt:}

\begin{lstlisting}
You will be given a list of reasoning segment nodes, each labeled
(e.g., N1, N2, ...), representing steps in a reasoning trace.

Your task has to strictly follow three steps:

First, analyze each node:
Carefully read all reasoning nodes and provide a brief analysis of each
node's function in the reasoning process.

Second, assign reasoning strategies:
Using the taxonomy provided below, identify the primary reasoning
strategy each node represents. Some longer nodes may involve multiple
strategies if so, also indicate a secondary strategy. If no secondary
applies, leave it as "None".

Finally, strictly format your output as JSONL as below:

{"id": "N1", "taxonomy_primary_type": "verification", "taxonomy_secondary_type": null}
{"id": "N2", "taxonomy_primary_type": "verification", "taxonomy_secondary_type": "Exploration"}

Reasoning Taxonomy

There are 5 types of reasoning strategies:

Backtracking:
The node revisits and modifies a previous step or assumption to correct
an error, resolve a conflict, or incorporate a new insight that alters
the reasoning path.

Verification:
The node tests or confirms the correctness of a specific claim,
assumption, or result without modifying the reasoning path.

Exploration:
The node proposes new hypotheses, possibilities, or approaches to the
problem in an open-ended manner, without committing to a definitive
solution.

Clarification:
The node rephrases, restates, or defines terms, assumptions, or problem
constraints to reduce ambiguity and enhance understanding.

Conclusion:
The node synthesizes prior reasoning to assert a final or intermediate
solution, judgment, or result.

Node:
["N1: <think>\nFirst, I recognize that the area of a circle is given by
the formula \( A = \pi r^2 \). Given that the area is \( 49\pi \), I can
set up the equation \( 49\pi = \pi r^2 \). To find the radius, I'll divide
both sides of the equation by \( \pi \), which simplifies to \( r^2 = 49 \).
Taking the square root of both sides, I get \( r = \sqrt{49} \), and since
the radius can't be negative, \( r = 7 \). Therefore, the radius of the
circle is 7 units."]
\end{lstlisting}

\textbf{Tree construction prompt:}

\begin{lstlisting}
You will be given a list of reasoning segment nodes, each labeled
(e.g., N1, N2, ...), representing steps in a mathematical reasoning
process.

Your task is to analyze the logical structure and organize these
segments step by step into an n-ary tree, where:

- Each node corresponds to one or more labels
  (e.g., "N1", or "N3, N4" if they share the same logical role).
- Labels must remain sequential. For example, N5 cannot be placed
  before N4.
- The tree reflects the hierarchical and sequential logic of the reasoning.
- Child nodes represent the logic going to the next level, such as
  supporting arguments, elaborations, or consequences.
- Sibling nodes represent logic staying at the same level.

Each time you add a node to the tree, you have three options:
A. Add as a sibling to one of the ancestor nodes of the current leaf node.
B. Add as a sibling to the current leaf node.
C. Add as a child of the current leaf node.

Provide your thought process and then give the tree.
Format the tree output as a structured JSON tree using only the node
labels do not include the original text.

Example tree format:
{
  "id": "1",
  "label": "N1",
  "children": [
    {
      "id": "1.1",
      "label": "N2",
      "children": [
        {"id": "1.1.1", "label": "N3"},
        {"id": "1.1.2", "label": "N4"}
      ]
    }
  ]
}

Nodes:
["N1: <think>\nFirst, I recognize that the area of a circle is given by
the formula \( A = \pi r^2 \). Given that the area is \( 49\pi \), I can
set up the equation \( 49\pi = \pi r^2 \). To find the radius, I'll divide
both sides of the equation by \( \pi \), which simplifies to \( r^2 = 49 \).
Taking the square root of both sides, I get \( r = \sqrt{49} \), and since
the radius can't be negative, \( r = 7 \). Therefore, the radius of the
circle is 7 units."]
\end{lstlisting}

\textbf{ECN conclusion judgment prompt:}

\begin{lstlisting}
Task:
You are given a reasoning trace divided into nodes (e.g., N1, N2, ...).
Each node represents a step in the reasoning process.

You are also given a list of conclusion nodes: ['N1'] identified by
their serial numbers.

For each conclusion node, do the following step by step:

1. Determine whether the conclusion node matches the Correct Answer.
   - If the conclusion in the supplied THINK TRACE for which the trace
already contains a correct, complete, logically sufficient solution to the QUESTION, mark it as 1.
   - If it does not match, mark it as 0.

2. Determine the type of the conclusion node:
   - Intermediate Conclusion: a step that supports later reasoning but
     does not directly answer the question.
   - Answering Conclusion: a conclusion contains a complete, logically sufficient solution to the QUESTION

3. Include the question and the Correct Answer:
   - Question: The area of a circle is \(49\pi\) square units. What is the
     radius of the circle, in units?
   - Correct Answer: 7

4. Export the result in JSONL format, including the node id,
   correctness, and type. Example:

{"conclusion_node": "N5", "is_correct": 0, "type": "Intermediate Conclusion"}
{"conclusion_node": "N10", "is_correct": 1, "type": "Answering Conclusion"}
{"conclusion_node": "N12", "is_correct": 1, "type": "Answering Conclusion"}

Complete Nodes:
["N1: <think>\nFirst, I recognize that the area of a circle is given by
the formula \( A = \pi r^2 \). Given that the area is \( 49\pi \), I can
set up the equation \( 49\pi = \pi r^2 \). To find the radius, I'll divide
both sides of the equation by \( \pi \), which simplifies to \( r^2 = 49 \).
Taking the square root of both sides, I get \( r = \sqrt{49} \), and since
the radius can't be negative, \( r = 7 \). Therefore, the radius of the
circle is 7 units."]
\end{lstlisting}

\clearpage
\section{Additional Experimental Results}
\label{app:additional results}

\subsection{Benchmark Results with Gemini Annotations}

In Table~\ref{tab:main_results} and Table~\ref{tab:deepseek_base_sd_comparison}, we observe that our findings are consistent across various annotation models.

\begin{table*}[!tbp]
\centering
\small
\setlength{\tabcolsep}{2.5pt}
\begin{tabular}{ll|cccc|cccc|cccc}
\toprule
& & \multicolumn{4}{c|}{\textbf{AIME'24}}
& \multicolumn{4}{c|}{\textbf{GSM8K}}
& \multicolumn{4}{c}{\textbf{Math 500}} \\
\cmidrule(lr){3-6}
\cmidrule(lr){7-10}
\cmidrule(lr){11-14}
\textbf{Model} & \textbf{Method}
& \textbf{Acc} & \textbf{$\Delta$} & \textbf{Tok} & \textbf{$\Delta$\%}
& \textbf{Acc} & \textbf{$\Delta$} & \textbf{Tok} & \textbf{$\Delta$\%}
& \textbf{Acc} & \textbf{$\Delta$} & \textbf{Tok} & \textbf{$\Delta$\%} \\
\midrule
\multirow{3}{*}{Qwen}
& Base
& .367 & -- & 10848 & --
& .901 & -- & 1311 & --
& .878 & -- & 3678 & -- \\
& SD-Full
& .467 & +.100 & 11038 & +1.7
& .901 & +.000 & 1478 & +12.7
& .852 & -.026 & 3321 & -9.7 \\
\rowcolor{gray!15}
& SD-ECN (ours)
& \textbf{.500} & \textbf{+.133} & \textbf{8694} & \textbf{-19.9}
& \textbf{.916} & \textbf{+.015} & \textbf{900} & \textbf{-31.3}
& \textbf{.886} & \textbf{+.008} & \textbf{2138} & \textbf{-41.9} \\
\midrule
\multirow{3}{*}{Llama}
& Base
& .433 & -- & 11855 & --
& .840 & -- & 1535 & --
& .879 & -- & 3413 & -- \\
& SD-Full
& .400 & -.033 & 11557 & -2.5
& .884 & +.044 & 1587 & +3.4
& \textbf{.844} & \textbf{-.035} & 3447 & +1.0 \\
\rowcolor{gray!15}
& SD-ECN (ours)
& \textbf{.533} & \textbf{+.100} & \textbf{8713} & \textbf{-26.5}
& \textbf{.893} & \textbf{+.053} & \textbf{933} & \textbf{-39.2}
& \textbf{.844} & \textbf{-.035} & \textbf{2256} & \textbf{-33.9} \\
\bottomrule
\end{tabular}
\caption{Reasoning accuracy and token efficiency of Base, SD-Full, and SD-ECN (ours) on AIME'24, GSM8K, and Math 500 using \texttt{DeepSeek-V4-Flash} annotation. SD denotes self-distillation. Acc denotes accuracy and Tok denotes token count. $\Delta$ is the absolute accuracy change relative to the corresponding Base model, and $\Delta$\% is the relative token change. \textbf{Bold} marks the better result between SD-Full and SD-ECN for each model and benchmark: higher accuracy and accuracy change, and lower token count and relative token change.}
\label{tab:deepseek_base_sd_comparison}
\end{table*}

\begin{figure}[!tbp]
    \centering
    \includegraphics[width=\linewidth]{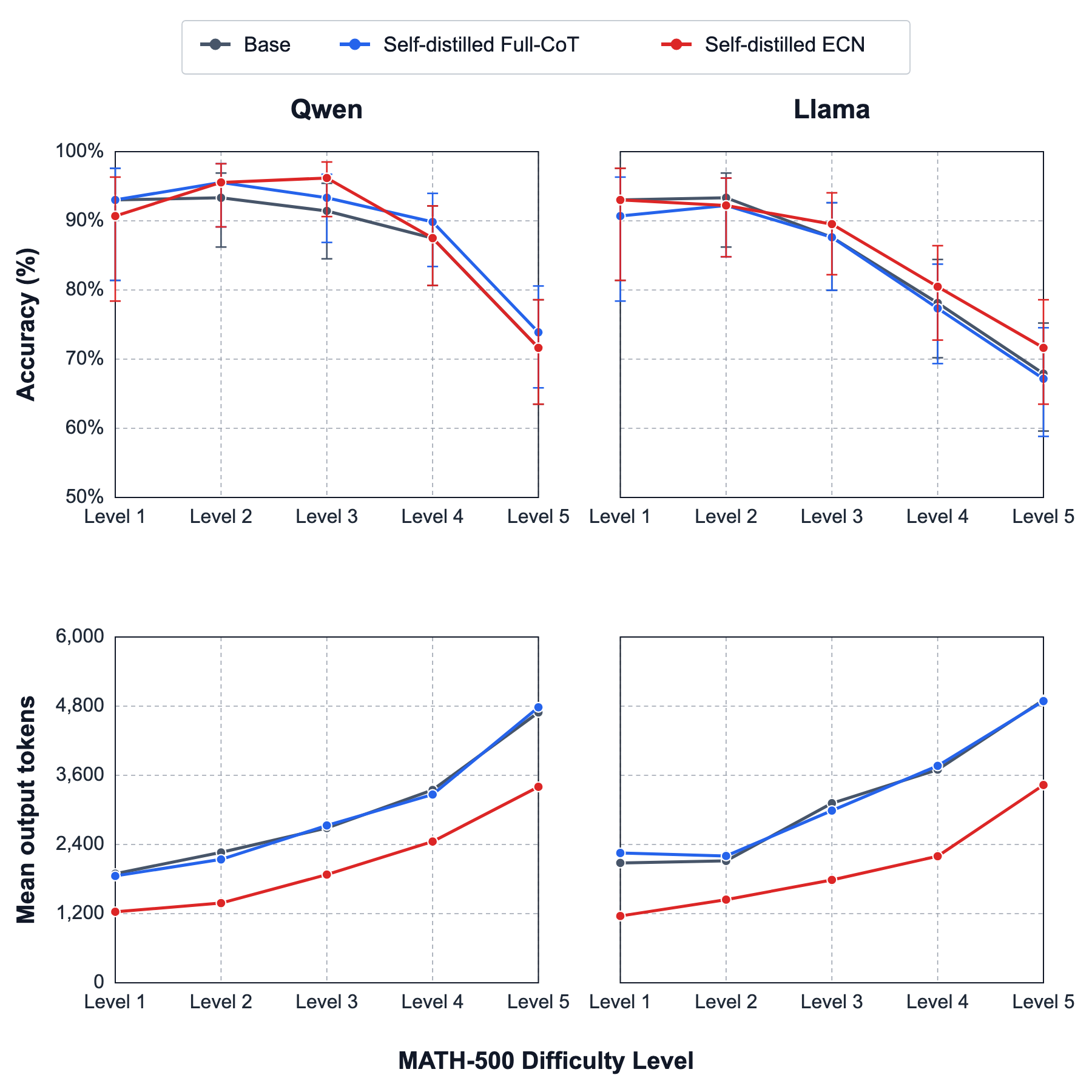}
    \caption{Accuracy and token usage across MATH-500 difficulty levels. The top row shows accuracy and the bottom row shows mean output tokens; columns correspond to Qwen and Llama. Self-distilled ECN maintains competitive accuracy across levels while reducing output length. Error bars denote 95\% Wilson confidence intervals.}
    \label{fig:math500_difficulty_combined}
\end{figure}

\paragraph{ECN performance across MATH-500 difficulty levels.}
Figure~\ref{fig:math500_difficulty_combined} examines whether ECN-based self-distillation remains effective as MATH-500 problems become more difficult. 1. Accuracy.
Across both Qwen and Llama, Self-distilled ECN exhibits a gradual decline in accuracy as difficulty increases from Level~1 to Level~5, following a trend similar to Base and Self-distilled Full-CoT. ECN remains competitive throughout this progression, without an abrupt accuracy drop at higher difficulty levels. Although its relative advantage varies across models and levels, its accuracy consistently tracks or exceeds the baselines as difficulty increases. These results support the stability of ECN across difficulty levels, rather than a uniform accuracy improvement over the baselines. 2.Token efficiency. Across both model families and all five difficulty levels, Self-distilled ECN uses fewer output tokens than both Base and Self-distilled Full-CoT. The reduction is largest on the easiest problems: relative to Base, ECN reduces output tokens by 49.9\% for Qwen and 54.1\% for Llama at Level~1. The reduction remains substantial, but is smaller, at Level~5 (28.2\% for Qwen and 24.7\% for Llama). This pattern is consistent with ECN removing redundant post-solution reasoning that is more prevalent on easier problems; harder problems require intrinsically longer solution processes, leaving less redundant reasoning to prune. Thus, ECN-based target construction reduces reasoning cost across difficulty levels while preserving competitive accuracy on difficult problems.

\subsection{Training Dynamics and Reasoning Trace Analysis}
\begin{figure}[htbp]
    \centering
    \includegraphics[width=0.78\linewidth]{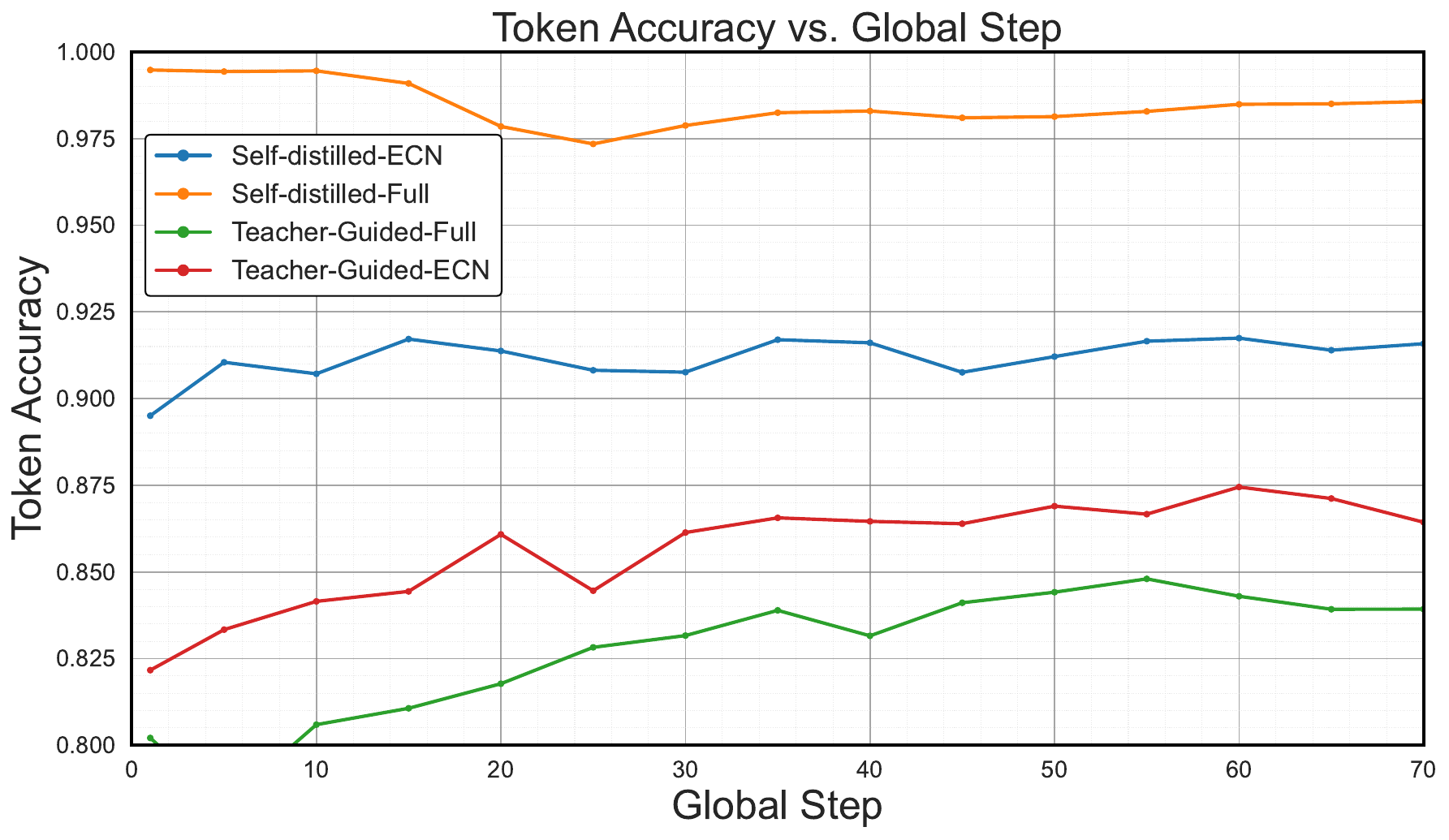}
    \caption{Token accuracy over training steps for different supervision and pruning strategies.}
    \label{fig:token_accuracy}
\end{figure}

The training dynamics show that this structural shift remains stable under self-distilled supervision. Figure~\ref{fig:token_accuracy} plots token accuracy over training steps for four supervision variants: Self-distilled-Full, Self-distilled-ECN, Teacher-Guided-Full, and Teacher-Guided-ECN. Self-distilled-Full maintains the highest token accuracy throughout, indicating strong alignment between the student model and its own traces. Applying ECN under self-distilled supervision lowers token accuracy relative to Self-distilled-Full, but the curve remains stable and gradually improves, showing that pruning changes the supervision signal without destabilizing optimization. In contrast, teacher-guided supervision yields lower token accuracy, consistent with a larger off-policy mismatch between teacher traces and the student model. This interpretation is reinforced by the token-level KL analysis in Figure~\ref{fig:kl_checkpoint}. 

We provide additional visualizations that complement the main results. Figure~\ref{fig:math500_token_scatter} compares the response-length distribution of the base Qwen model and the self-distilled ECN-Qwen model on Math~500 (Level~3). Figure~\ref{fig:token_accuracy} shows token accuracy over training steps for different supervision and pruning strategies. Figure~\ref{fig:self-distilled-low-data-deltas} compares self-distilled and teacher-guided training under different supervision budgets.

\begin{table}[htbp]
\centering
\small
\setlength{\tabcolsep}{4pt}
\begin{tabular}{lrrr}
\toprule
Metric & SD-ECN & Base & $\Delta$ \\
\midrule
Earliest final-answer depth & 3.181 & 3.029 & +0.152 \\
Post-answer source-node gap & 5.762 & 12.143 & -6.381 \\
\bottomrule
\end{tabular}
\caption{Gemini Annotation: Answer-emergence timing and post-answer continuation on 105 Math~500 Level~3 traces. Lower post-answer gap indicates less redundant continuation.}
\label{tab:search-pruning}
\end{table}

\paragraph{Annotation cut positions.}
We further examine annotation-selected cut positions on 865 shared Qwen reasoning traces (Table~\ref{tab:annotation_cut_positions}). DeepSeek retains a larger fraction of the source reasoning than Gemini (61.79\% vs.\ 56.84\%), but its retained-prefix distribution has a lower median and a higher 90th percentile. Under a common heuristic segmentation, the two constructions select the same cut node on 40.35\% of traces. These results show that the annotation procedure changes the distribution of retained reasoning, rather than simply shifting all cutoffs to later positions.

\begin{table}[t]
\centering
\small
\setlength{\tabcolsep}{4pt}
\begin{tabular}{@{}lrr@{}}
\toprule
Statistic & Gemini & DeepSeek \\
\midrule
Retained CoT (\%) & 56.84 & 61.79 \\
Mean cut position (chars) & 5,656 & 6,148 \\
Median cut position (chars) & 3,902 & 3,496 \\
Full trace retained (\%) & 12.83 & 5.32 \\
\bottomrule
\end{tabular}
\caption{
Cut-position statistics on 865 shared Qwen Full-CoT traces.
Positions are measured as retained source-prefix characters, not tokens.
Retained CoT is the ratio of aggregate retained to source characters;
}
\label{tab:annotation_cut_positions}
\end{table}

\subsection{Sensitivity to the Annotation Model}
\label{sec:annotation-model-sensitivity}

\begin{figure}[!t]
    \centering
    \includegraphics[width=\columnwidth]{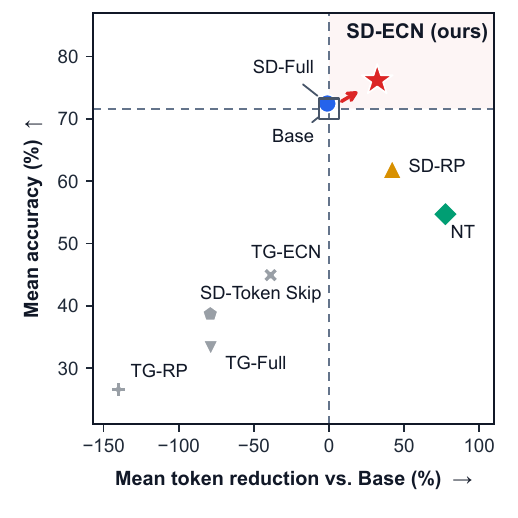}
    \caption{
        \textbf{Aggregate accuracy--token trade-off with DeepSeek annotations.}
        Results are averaged equally over two backbones and three benchmarks.
        Token changes are computed relative to the corresponding Base model
        before averaging. Dashed lines indicate Base; upper right is better.
    }
    \label{fig:deepseek-annotation-tradeoff}
\end{figure}

We repeat the ECN annotation and target-construction pipeline using
\texttt{DeepSeek-V4-Flash}. As shown in
Figure~\ref{fig:deepseek-annotation-tradeoff}, SD-ECN improves mean
accuracy from 71.6\% to 76.2\% while reducing output-token usage by
32.1\% relative to Base. In comparison, SD-Full reaches 72.5\% mean
accuracy and increases token usage by 1.1\%. These results indicate that
ECN's accuracy and efficiency benefits are not specific to a single
annotation model.

The DeepSeek-annotated results provide further evidence that the benefits
of ECN are not specific to a single annotation model. Under Gemini
annotations, SD-ECN achieves 72.6\% mean accuracy with a 32.5\% reduction
in output-token usage relative to Base. With DeepSeek annotations, it
achieves 76.2\% mean accuracy while maintaining a comparable 32.1\%
token reduction. The efficiency benefit therefore remains consistent
across the two annotation settings, while the aggregate accuracy gain
over Base increases from 1.0 to 4.6 percentage points.

By comparison, SD-Full reaches 72.5\% mean accuracy and increases
output-token usage by 1.1\% on average in the DeepSeek-annotated
evaluation. Thus, in this setting, SD-ECN achieves both higher aggregate
accuracy and substantially lower output-token usage than full-chain
self-distillation. At the benchmark level, SD-ECN improves over Base in
five of the six model--benchmark settings while reducing output-token
usage in all six.

\section{Human Annotation Evaluation}
\label{sec:human-annotation}

To evaluate the reliability of the model-assisted annotation pipeline,
we conduct a human evaluation comparing LLM-generated annotations
with human annotations for two key tasks: taxonomy labeling and
conclusion correctness judgment.

\subsection{Annotation Protocol}

To evaluate the reliability of the model-assisted annotation pipeline,
we conduct a human evaluation on a randomly sampled subset of the
training data. Specifically, we randomly sample 20 reasoning traces
from the PRM-12k training set. After segmentation, these traces contain
a total of 463 reasoning nodes, each associated with a taxonomy label
predicted by the LLM.

One human annotator manually reviews and labels the sampled nodes using
the same annotation guidelines described in Appendix~\ref{app:prompts}.
For each node, the annotator assigns the primary taxonomy label
(\textit{Backtracking}, \textit{Verification}, \textit{Exploration},
\textit{Clarification}, or \textit{Conclusion}). For nodes identified
as conclusions, the annotator additionally determines whether the node
corresponds to the correct answer and whether it should be categorized
as an \textit{Intermediate Conclusion} or an \textit{Answering Conclusion}.

\subsection{Taxonomy Labeling Accuracy}

We evaluate taxonomy labeling by measuring the agreement between the
primary taxonomy labels generated by Gemini-2.5-Flash and the human
annotations. On the sampled 463 reasoning nodes, the agreement rate
between the LLM annotations and the human labels is 88.9\%, indicating
a high level of consistency.

Most disagreements arise from the heuristic segmentation procedure.
In some cases, a single reasoning segment may simultaneously exhibit
multiple reasoning functions (e.g., \textit{Exploration} together with
an \textit{Intermediate Conclusion}). When such mixed reasoning occurs,
the human annotator and the LLM may assign different primary labels,
even though both labels are reasonable interpretations of the segment.
Therefore, these cases should not necessarily be interpreted as clear
annotation errors.

\subsection{Conclusion Judgment Accuracy}

We further evaluate the reliability of the ECN conclusion judgment
step. For this evaluation, we randomly sample 50 reasoning traces and
compare the LLM predictions with human annotations.

Two aspects are evaluated: (1) \textit{correctness detection}, which
determines whether a predicted conclusion matches the correct answer,
and (2) \textit{first-conclusion identification}, which checks whether
the earliest answering conclusion predicted by the LLM matches the one
identified by the human annotator.

On the sampled traces, the correctness detection accuracy reaches
100\%, and the first-conclusion identification accuracy also reaches
100\%. These results suggest that the LLM-based annotation pipeline
provides reliable signals for identifying the first correct answering
conclusion used in our pruning method.

\section{Metric Definitions}
\label{sec:metric_definitions}

This section defines the trace-level metrics used in our analysis. Each reasoning
trace is first represented as a grouped tree, while the original source reasoning
segments are retained as an ordered sequence. We distinguish between metrics
computed on the grouped tree structure and metrics computed from the source-node
order.

\subsection{Notation}

We model each reasoning trace as a rooted \(n\)-ary tree
\[
T=(V,E),
\]
where \(V\) is the set of grouped tree nodes, \(E\) is the set of parent-child
edges, and the root has depth \(0\). Depth is measured in edges.

For any node \(v\in V\), let \(d(v)\) denote its depth and let
\[
b(v)=|\mathrm{child}(v)|
\]
denote its branching factor. We define the leaf set as
\[
\mathcal{L}=\{\ell\in V:\mathrm{child}(\ell)=\varnothing\},
\]
and the internal-node set as
\[
\mathcal{I}=\{v\in V:|\mathrm{child}(v)|>0\}.
\]

Each grouped tree node may contain one or more original source reasoning
segments. Let
\[
S=(N_1,N_2,\dots,N_m)
\]
be the ordered source-node sequence for a trace. For each source node \(N_i\),
let \(g(N_i)\in V\) denote the grouped tree node that contains it.

Unless otherwise specified, variance terms are computed as population variances.
For degenerate cases, such as a single-node tree with no internal nodes, the
corresponding internal-node metrics are set to zero.

\subsection{Tree-Node Metrics}
\label{app:structural-metrics}

The metrics in this subsection are computed on the grouped tree \(T\).

\paragraph{Total Tree Nodes.}
The total number of grouped tree nodes is
\[
\mathrm{TotalTreeNodes}(T)=|V|.
\]

\paragraph{Maximum Depth.}
The maximum root-to-leaf depth is
\[
\mathrm{MaxDepth}(T)=\max_{\ell\in\mathcal{L}} d(\ell).
\]

\paragraph{Maximum Width.}
For each depth \(h\), let
\[
W_h=|\{v\in V:d(v)=h\}|
\]
be the number of grouped tree nodes at that depth. The maximum width is
\[
\mathrm{MaxWidth}(T)=\max_h W_h.
\]

\paragraph{Width-to-Depth Ratio.}
The width-to-depth ratio compares the widest layer of the tree with its maximum
depth:
\[
\mathrm{WDR}(T)=\frac{\mathrm{MaxWidth}(T)}{\mathrm{MaxDepth}(T)}.
\]
For single-node trees, where \(\mathrm{MaxDepth}(T)=0\), we set
\(\mathrm{WDR}(T)=0\).

\paragraph{Mean Leaf Depth.}
The mean depth of leaf nodes is
\[
\mathrm{MeanLeafDepth}(T)
=\frac{1}{|\mathcal{L}|}\sum_{\ell\in\mathcal{L}} d(\ell).
\]

\paragraph{Leaf-Depth Variance.}
The variance of leaf depths is
\[
\begin{aligned}
&\mathrm{LeafDepthVar}(T) \\
&\quad=\frac{1}{|\mathcal{L}|}\sum_{\ell\in\mathcal{L}}
\left(d(\ell)-\mathrm{MeanLeafDepth}(T)\right)^2.
\end{aligned}
\]

\paragraph{Mean Branching Factor.}
The mean branching factor over internal nodes is
\[
\mathrm{MeanBranching}(T)
=\frac{1}{|\mathcal{I}|}\sum_{v\in\mathcal{I}} b(v).
\]

\paragraph{Maximum Branching Factor.}
The maximum branching factor over internal nodes is
\[
\mathrm{MaxBranching}(T)=\max_{v\in\mathcal{I}} b(v).
\]

\paragraph{Branching Variance.}
The variance of internal-node branching factors is
\[
\begin{aligned}
&\mathrm{BranchingVar}(T) \\
&\quad=\frac{1}{|\mathcal{I}|}\sum_{v\in\mathcal{I}}
\left(b(v)-\mathrm{MeanBranching}(T)\right)^2.
\end{aligned}
\]

\paragraph{Structural Efficiency.}
Let
\[
\begin{aligned}
&\mathrm{LongestPathNodes}(T) \\
&\quad=\max_{\ell\in\mathcal{L}} |P(r,\ell)|
\end{aligned}
\]
denote the maximum root-to-leaf path length counted in nodes rather than edges,
where \(P(r,\ell)\) is the set of nodes on the path from the root \(r\) to leaf
\(\ell\). We define
\[
\begin{aligned}
&\mathrm{StructuralEfficiency}(T) \\
&\quad=\frac{\mathrm{LongestPathNodes}(T)}{\mathrm{TotalTreeNodes}(T)}.
\end{aligned}
\]
This quantity measures the fraction of grouped tree nodes covered by the
longest reasoning path.

\paragraph{Sackin Index.}
The Sackin index is the sum of leaf depths:
\[
\mathrm{Sackin}(T)=\sum_{\ell\in\mathcal{L}} d(\ell).
\]

\paragraph{Generalized Colless Index.}
For any internal node \(v\), let \(L(u)\) be the number of descendant leaves
under child \(u\in\mathrm{child}(v)\). We define the local imbalance at \(v\) as
\[
\mathrm{LocalImbalance}(v)
=
\sum_{\substack{u_i,u_j\in\mathrm{child}(v)\\ i<j}}
|L(u_i)-L(u_j)|.
\]
The generalized Colless index is then
\[
\mathrm{GenColless}(T)
=
\sum_{v\in\mathcal{I}} \mathrm{LocalImbalance}(v).
\]
Larger values indicate greater imbalance among sibling subtrees.

\subsection{Source-Node Metrics}

The metrics in this subsection are defined from the ordered source-node sequence
\(S\). When depth information is needed, the corresponding source node is mapped
back to its grouped tree node through \(g(\cdot)\).

\paragraph{Final-Answer Proxy.}
Let \(A^\star\) be the final-answer proxy extracted from the latest source node
that contains either a boxed answer or the literal string \texttt{Final Answer}.
If no such proxy is found, the trace is excluded from the answer-boundary
metrics.

\paragraph{Earliest Final-Answer Node.}
Given \(A^\star\), let
\[
k^\star
=
\min\{i\in\{1,\dots,m\}: A^\star \text{ appears in } N_i\}.
\]
The earliest final-answer node is
\[
\mathrm{EarliestFinalAnswerNode}(S)=N_{k^\star}.
\]

\paragraph{Earliest Final-Answer Depth.}
After locating \(N_{k^\star}\) in source-node order, we map it to the grouped
tree and report its depth:
\[
\mathrm{EarliestFinalAnswerDepth}(S,T)
=
d\!\left(g(N_{k^\star})\right).
\]
Thus, the answer boundary is detected in source-node order and then evaluated
with respect to the grouped-tree structure.

\paragraph{Post-answer Source-Node Gap.}
The remaining source-node distance after the earliest final-answer node is
\[
\mathrm{PostAnswerSourceGap}(S)
=
m-k^\star.
\]
Equivalently, this is the number of source nodes that appear after the earliest
occurrence of the final-answer proxy.

\end{document}